\documentclass[letterpaper]{article} 
\usepackage{arxiv_aaai25}  
\usepackage{times}  
\usepackage{helvet}  
\usepackage{courier}  
\usepackage[hyphens]{url}  
\usepackage{graphicx} 
\usepackage[numbers]{natbib}  
\usepackage{caption} 
\usepackage{algorithm}
\usepackage{algorithmic}

\usepackage{newfloat}
\usepackage{listings}
\DeclareCaptionStyle{ruled}{labelfont=normalfont,labelsep=colon,strut=off} 
\floatstyle{ruled}
\newfloat{listing}{tb}{lst}{}
\floatname{listing}{Listing}
\usepackage{hyperref}
\hypersetup{
	colorlinks=true,
	linkcolor=cyan,
	filecolor=blue,      
	urlcolor=red,
	citecolor=green,
}
\usepackage{multirow}
\usepackage{multicol}
\usepackage{subfig}
\usepackage{amssymb}
\usepackage{xcolor}
\usepackage{amsmath}
\usepackage{array}

\newcolumntype{L}[1]{>{\raggedright\arraybackslash}p{#1}}
\newcolumntype{C}[1]{>{\centering\arraybackslash}p{#1}}
\newcolumntype{R}[1]{>{\raggedleft\arraybackslash}p{#1}}

\renewcommand{\thefootnote}{\fnsymbol{footnote}}

\title{BearLLM: A Prior Knowledge-Enhanced Bearing Health Management Framework with Unified Vibration Signal Representation}
\author{
    Haotian Peng\textsuperscript{\rm 1, 2, 3, 4}\equalcontrib,
    Jiawei Liu\textsuperscript{\rm 1, 2, 3}\equalcontrib,
    Jinsong Du\textsuperscript{\rm 1, 2 ,3},
    Jie Gao\textsuperscript{\rm 1, 2, 3}\footnotemark[2],
    Wei Wang\textsuperscript{\rm 1, 2, 3}\footnotemark[2]
}
\affiliations{
    \textsuperscript{\rm 1} Shenyang Institute of Automation, Chinese Academy of Sciences\\
    \textsuperscript{\rm 2} Liaoning Liaohe Laboratory 
    \textsuperscript{\rm 3} Key Laboratory on Intelligent Detection and Equipment Technology of Liaoning Province\\
    \textsuperscript{\rm 4} University of Chinese Academy of Sciences\\
    {\{penghaotian, liujiawei, jsdu, gaojie, wangwei2\}@sia.cn}

}

\usepackage{bibentry}

\begin{document}

\maketitle

\begin{abstract}

    We propose a bearing health management framework leveraging large language models (BearLLM), a novel multimodal model that unifies multiple bearing-related tasks by processing user prompts and vibration signals. Specifically, we introduce a prior knowledge-enhanced unified vibration signal representation to handle various working conditions across multiple datasets. This involves adaptively sampling the vibration signals based on the sampling rate of the sensor, incorporating the frequency domain to unify input dimensions, and using a fault-free reference signal as an auxiliary input. To extract features from vibration signals, we first train a fault classification network, then convert and align the extracted features into word embedding, and finally concatenate these with text embedding as input to an LLM. To evaluate the performance of the proposed method, we constructed the first large-scale multimodal bearing health management (MBHM) dataset, including paired vibration signals and textual descriptions. With our unified vibration signal representation, BearLLM using one set of pre-trained weights achieves state-of-the-art performance on nine publicly available fault diagnosis benchmarks, outperforming specific methods designed for individual datasets. We provide a dataset, our model, and code to inspire future research on building more capable industrial multimodal models (\href{https://github.com/SIA-IDE/BearLLM}{https://github.com/SIA-IDE/BearLLM}).

\end{abstract}

\section{Introduction}
\renewcommand{\thefootnote}{\fnsymbol{footnote}}
\footnotetext[2]{Corresponding author.}
\renewcommand{\thefootnote}{\arabic{footnote}}

Bearings are the core components of mechanical rotating equipment but have high failure rates due to complex operational and environmental conditions \cite{Wang2020IntelligentBearing}. Bearing health management (e.g., anomaly detection, fault diagnosis, and maintenance recommendations) is of great practical significance in industrial safety production to reduce economic losses and maintenance costs \cite{pengDigitalTwinRolling2022, xiaoNovelJointTransfer2022a, ruanCNNParameterDesign2023a}.

\begin{figure}
    \centering
    \includegraphics[width=3.3in]{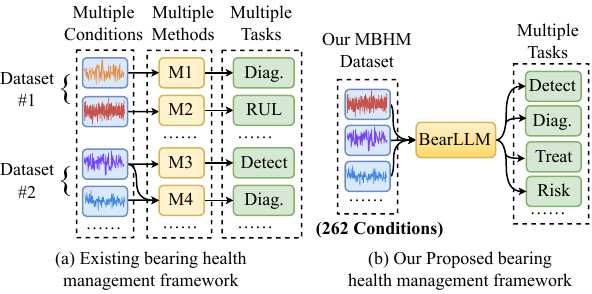}
    \caption{Comparison of existing bearing health management frameworks \cite{chaleshtoriNovelBearingFault2024,niDatadrivenBearingHealth2024} with our proposed approach. Our BearLLM replaces the complex operations of designing methods tailored to different conditions and tasks.}
    \label{fig:framework_compare}
    \vspace{-0.2cm}
\end{figure}

Current bearing health management frameworks rely on designing specialized methods for different working conditions and tasks, as shown in Fig. \ref{fig:framework_compare} (a).
To apply specific methods to complex real-world industrial scenarios, domain adaptation, and generalization have attracted widespread attention. Domain adaptation enables a model trained on one source domain to perform well on different but related target domains by reducing the domain shift or discrepancy \cite{wuAdversarialDomainAdaptation2022, zhangSupervisedContrastiveLearningBased2022}, but it suffers from low accuracy when the source and target domains are category-inconsistent (e.g., transitioning from working condition $C_1$ with four fault types to $C_2$ with five types). Domain generalization aims to extract domain-invariant features to improve performance on unseen domain \cite{liDomainGeneralizationRotating2020, zhengDeepDomainGeneralization2021, chenAdversarialDomainInvariantGeneralization2022a}, but it is often constrained to a limited number of working conditions with small differences, e.g., fewer than ten working conditions in \cite{chenAdversarialDomainInvariantGeneralization2022, linGeneralizedMAMLFewshot2023}. 
These purely data-driven methods often fail to strike an optimal balance between high accuracy and strong generalization for fault diagnosis.

In this paper, we propose a prior knowledge-enhanced bearing large language model (BearLLM), which can unify multiple bearing health management tasks over hundreds of different working conditions from multiple datasets, as shown in Fig. \ref{fig:framework_compare} (b). 
To handle various working conditions, we introduce a prior knowledge-enhanced unified vibration signal representation. Unlike most fault diagnosis methods that use fixed-length input segments, we sample vibration signals as variable-length but fixed-duration segments. These duration-consistent segments are then converted to the frequency domain and are aligned. We further utilize a fault-free reference signal as a prior input, eliminating the need for complex mechanism analysis for various bearing designations \cite{zhengDeepDomainGeneralization2021}.

Specifically, we first design a fault classification network (FCN) to extract fault features based on the differences in frequency components between the query signal segment and the fault-free reference signal segment. 
This new frequency-based feature extraction paradigm for bearing fault diagnosis is more efficient (i.e., faster convergence and higher accuracy) and achieves stronger generalization, compared to previous methods that extract fault features directly from vibration signals.
The extracted features are then transformed and aligned into word embedding, which is subsequently connected to user text embedding as inputs to the LLM.
To evaluate the performance of the proposed method, we construct the first large-scale multimodal bearing health management (MBHM) dataset, including paired vibration signals and textual descriptions. Although the vibration signals from the nine public datasets differ significantly in distribution, BearLLM with a set of pre-trained weights achieves state-of-the-art performance using a unified vibration signal representation, outperforming specialized methods designed for individual datasets. The contributions of this paper are summarized as follows:
\begin{itemize}
    \item We propose a novel bearing multimodal large language model, unifying multiple bearing health management tasks by aligning vibration signals and textual prompts.
    \item We propose a prior knowledge-enhanced unified vibration signal representation to handle various working conditions from multiple datasets.
    \item We construct the first large-scale multimodal dataset for bearing health management (MBHM), involving vibration signals with associated textual descriptions. 
    \item  Experimental results show that our BearLLM outperforms state-of-the-art fault diagnosis methods on nine publicly available benchmarks.  
\end{itemize}

\begin{figure*}
    \centering
    \includegraphics[width=7in]{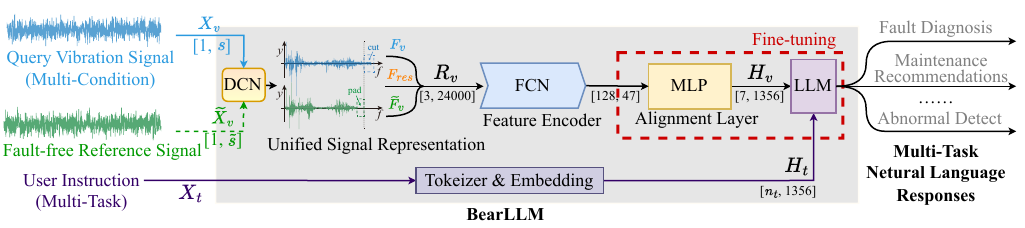}
    \caption{Architecture of our proposed BearLLM. Given a query vibration signal segment $X_v$ and user instruction $X_t$ as input, the model retrieves a fault-free vibration signal segment $\tilde{X}_v$ with similar working conditions from the database as a reference. Two vibration signals are converted into a unified representation through DCN. A feature encoder identifies fault-related residuals between the two signals. The alignment layer converts these features into the word embedding $H_V$. Finally, an LLM is utilized with the user text embedding $H_T$ to generate multi-task natural language responses, where $n_t$ represents the length of the encoded text embedding.}
    \label{fig:framework}
    \vspace{-0.2cm}
\end{figure*}

\section{Related Works}

\textbf{Multiple Working Condition:} Fault diagnosis under various working conditions from multiple datasets presents a challenge due to the heterogeneity of collected signals arising from variations in the test rigs, sensors, and environment, making it difficult to obtain unified features \cite{wenNovelDeepClustering2023}. Existing domain adaptation methods\cite{TCNN, wanNovelDeepConvolution2022,  maDigitalTwinassistedEnhanced2023, huoEnhancedTransferLearning2023} typically involves training a model under known working conditions (source domain) and subsequently transferring knowledge to an unknown working condition (target domain). However, these approaches still necessitate individual transfer fine-tuning for each working condition in practice, hindering its ability to generalize across multiple scenarios. Domain generalization methods leverage training on multiple working conditions and aim to align the feature distributions of different domains through the design of network architectures and loss functions \cite{jiaDeepCausalFactorization2023, zhuDomainAdaptationMultiAdversarial2023, houDiagnosisformerEfficientRolling2023}. However, these approaches often rely on complex data preprocessing and augmentation techniques to help models learn fault features from vibration signals.

\textbf{Multiple Tasks:} Data-driven machinery health management have gained significant traction \cite{trabelsiFMECABasedRiskAssessment2020}.The concept of health management usually involves multiple tasks \cite{omriAdaptedPHMApproach2021, zioPrognosticsHealthManagement2022}, including anomaly detection, fault diagnosis, degradation prediction, maintenance decision-making, etc. LLMs such as ChatGPT-4 \cite{openaiGPT4TechnicalReport2024} have demonstrated exceptional capabilities across a wide range of tasks. The emergence of open-source foundational models like LLaMA 3 \cite{llama3modelcard1} and Qwen 2 \cite{qwen} have further empowered researchers in various disciplines to integrate these models into their own applications. In the aviation domain, \citet{liuJointKnowledgeGraph2024} applied generalized linear models to achieve multiple tasks, including assembly guidance and assembly error identification for aircraft engines. In the petroleum industry, \citet{eckrothAnsweringNaturalLanguage2023} designed a question-answering system based on LLM and knowledge graph, enabling retrieval of functionalities such as stratigraphy data and geological age determination. However, research integrating multiple tasks using LLMs for bearing health management remains limited \cite{liChatGPTlikeLargescaleFoundation2024}. 

\section{A Multimodal Bearing Health Management Dataset}

Although several bearing-related datasets in Tab. \ref{tab:dataset_compare} are available, they generally collect vibration signals on a single test rig, have a limited number of working conditions, and have no corresponding textual descriptions for training LLM. We have constructed a large-scale publicly multimodal dataset for bearing health management (MBHM).

\begin{table}[H]
\small
    \centering
    \begin{tabular}{L{0.95cm}C{1.55cm}R{0.9cm}R{0.8cm}R{1.05cm}R{0.6cm}}
    \hline
        Dataset & Sample Rate (kHz) & Condi-tions$^{*}$ & Fault Types & Time (s) & Text \\ \hline
        CWRU & 12 / 48 & 12 & \textbf{10} & 3932 & $\times$  \\ 
        DIRG & 51.2 & 102 & 7 & 7140 & $\times$  \\ 
        HIT & 20 & 40 & 3 & 9648 & $\times$  \\ 
        IMS & 20 & 16 & 7 & \underline{46480} & $\times$  \\ 
        JNU & 100 & 45 & 4 & 3600 & $\times$  \\ 
        JUST & 50 & 36 & 4 & \underline{43986} & $\times$  \\ 
        MFPT & 48.8 / 97.6 & 1 & 3 & 78 & $\times$ \\ 
        PU & 64 & 4 & 5 & 7316 & $\times$  \\ 
        XJTU & 25.6 & 6 & \textbf{10} & 13336 & $\times$  \\ 
        \textbf{MBHM} & \textbf{12 \textasciitilde 100} & \textbf{262} & \textbf{10} & \textbf{135516} & \textbf{\checkmark} \\ \hline
        \multicolumn{6}{r}{$^{*}$ \scriptsize The same working conditions represent the same load, speed, and sensor.}\\
    \end{tabular}
    \caption{Comparison of different datasets. Our MBHM dataset has the largest number of working conditions, the most complete fault types, and the longest time, paired textual prompts/responses.}
    \label{tab:dataset_compare}
    \vspace{-0.2cm}
\end{table}

\begin{figure}[t]
    \centering
    \includegraphics[width=3.3in]{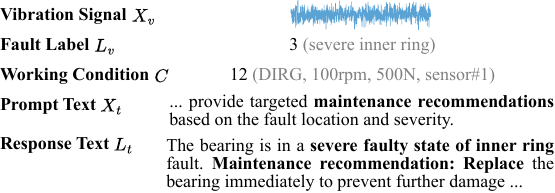}
    \caption{Sample case of our MBHM dataset, includes vibration signal $X_v$, fault label $L_v$, working condition $C$, the specific task prompt text $X_t$, and the response text $L_t$.}
    \label{fig:dataset}
    \vspace{-0.2cm}
\end{figure}
The MBHM contains 135,516 pairs of vibration signal segments and fault types, and 542,064 pairs of text cues and responses, of which each sample is shown in Fig. \ref{fig:dataset}, contains a vibration signal, a fault label, an operating condition id, a user prompt, and a text response, ie, $(X_v, L_v, C, X_t, L_t)\in \mathrm{MBHM}$.
Our dataset contains 262 working conditions collected from nine publicly accessible datasets, i.e., CWRU \cite{CWRU1}, DIRG \cite{DIRG}, HIT \cite{HIT}, IMS \cite{IMS}, JNU \cite{JNU1}, JUST \cite{JUST}, MFPT \cite{MFPT1}, PU \cite{PU}, XJTU \cite{XJTU}. For each vibration signal, we have four different tasks, i.e., anomaly detection, fault diagnosis, maintenance recommendations, and potential risk analysis by generating text responses using ChatGPT \cite{openaiGPT4TechnicalReport2024}. Detailed methodologies for dataset construction are provided in Appendix A.3. Our MBHM dataset contains the following features:
\begin{itemize}
    \item \textbf{Multi-modal:} Each vibration signal is paired with four text prompts and responses, supporting the training and development of multimodal multi-task models.
    \item  \textbf{Multiple working conditions:} Our dataset covers a wider range of working conditions, more accurately modeling real-world industrial production scenarios.
\end{itemize}

\section{Method}


In this section, we propose BearLLM, a novel multimodal model that unifies multiple bearing-related tasks. To handle various working conditions across multiple datasets, we introduce a prior knowledge-enhanced unified vibration signal representation in Section 4.1. The unified vibration signal is fed to a fault classification network to extract features in Section 4.2. We convert and align the extracted features into word embedding, and finally concatenate these with text embedding as input to an LLM in Section 4.3.

\subsection{Prior Knowledge-Enhanced Unified Vibration Signal Representation}
BearLLM aims to manage multiple bearing-related tasks across hundreds of working conditions. 
The basis for this is to build a unified vibration signal representation, involving adaptively sampling the vibration signal segments based on the sensor sampling rate, incorporating the frequency domain to unify input dimensions, and using a fault-free reference signal to calculate residual as auxiliary inputs to improve data utilization efficiency. 


\subsubsection{Adaptive Sampling}

To monitor various mechanical devices across different working conditions and industrial scenarios, vibration sensors are deployed with varying designations and sampling rates. However, most fault diagnosis methods \cite{zhuDomainAdaptationMultiAdversarial2023,dongRollingBearingIntelligent2024} use fixed-length signal segments in the time domain as inputs, where the fault frequency components in the inputs deviate from their original intrinsic values and vary with the sampling rate, hindering accurate fault diagnosis.
Instead of sampling fixed-length signal segments, we adaptively sample vibration signals as variable-length but fixed-duration segments using prior knowledge of the sensor sampling rate.
We extract the $m$-th query signal segment $X_v \in \mathbb{R}^{1 \times s}$ from the original signal $X_o$ by
\begin{equation}
    X_v = X_o[ms, (m+1)s],
    \label{eq:sampling}
\end{equation}
where $s$ denotes the sampling rate of the sensor and controls the length of the $X_v$. 

\subsubsection{Frequency-domain Input Alignment} 
After adaptive sampling, each query segment ($X_v$) has an equal duration, and the frequencies of $X_v$ are aligned. However, varying lengths of $X_v$ (due to different sampling rates) result in different numbers of frequency components, making them unsuitable for input to the network.
We design a discrete cosine normalization (DCN) that consists of converting the vibration signal to the frequency domain using the discrete cosine transform (DCT), unifying the number $n_f$ of frequency components using a pad or cut, and standardizing the amplitude using the normalization $\mathcal{N}$. The normalized frequency representation $F_v \in \mathbb{R}^{1 \times n_f}$ is obtained by
\begin{align}
F_v & = \left\{\begin{matrix}
  \mathcal{N}(\mathrm{DCT}(X_v)[0, n_f]), & \mathrm{if} \: s\ge n_f \\
  \mathcal{N}(\mathrm{DCT}(X_v)\cup [0]_{n_f-s}), & \mathrm{if} \: s< n_f
\end{matrix}\right.
\label{eq:prepocess_signal}
\end{align}
Signals with sampling rates below $n_f$ are zero-padded, while those exceeding $n_f$ are cut. To balance computational resources and fault classification accuracy, we empirically set $n_f = 24000$ (more detail in Tab. \ref{tab:target_length}).
To enhance training stability, the amplitude of the frequency sequence is normalized to $[-1,1]$,
\begin{equation}
    \mathcal{N}(x)=\beta\frac{\sqrt{n}x}{\|x\|_2},
    \label{eq:normalize_signal}
\end{equation}
where $\beta$ is a scaling factor and is set to 0.01 by statistically analyzing our MBHM dataset. 


\subsubsection{Fault-free Reference Signal}
To eliminate distributional differences from different inputs under various operating conditions, we introduce fault-free signals as reference signals.
1) In practical use, the reference signal segment $\tilde{X}_v$ can be collected and saved when the equipment is working properly, such as after factory acceptance or maintenance; 2) in training on the MBHM dataset, $\tilde{X}_v$ is acquired by
\begin{equation}
    \begin{aligned}
        \tilde{X}_v &\sim \left \{X_v^*|(X_v^*,L_v^*,C^*,X_t^*,L_t^*)\in \mathrm{MBHM}, \right. \\
        &\quad \left. L_v^*=0, C^*=C \right \}.
    \end{aligned}
    \label{eq:choice_normal}
\end{equation}
This indicates that $\tilde{X}_v$ is selected when a signal ($X_v^*$) of our MBHM dataset is fault-free (i.e., $L_v^*=0$) and has the same working conditions as ${X}_v$ (i.e., $C^*=C$).

We combine the query frequency signal ($F_v$), the fault-free frequency signal ($\tilde{F}_v$), and the residual frequency signal ($F_{res}=F_v-\tilde{F}_v$) as unified vibration signal representation,
\begin{equation}
    R_v=[F_v,\tilde{F}_v, F_{res}].
    \label{eq:input_sample}
\end{equation}


\subsection{Feature extraction}

\begin{algorithm}
    \caption{Training Algorithm}
    \label{algo:training}
    \begin{algorithmic}
    \REQUIRE $\theta_E, \theta_C, \theta_A, \theta_L$ the weights of feature encoder, linear classification layer, alignment layer, and LLM in our BearLLM; $X_v, L_v, X_t, L_{t}$ the vibration signal, fault label, prompt text, and response text from MBHM dataset 
    \ENSURE The optimal parameters of BearLLM $\theta_E^*, \theta_A^*, \theta_L^*$
        
        \STATE \textbf{Step 1: Pre-training FCN}
        \FOR{$e \leftarrow 1$ to 50 epoches}
        \STATE $R_v \leftarrow X_v$
        $\quad\triangleleft$
        get unified representation by Eq. \ref{eq:input_sample}
        \STATE $P = \mathrm{FCN}(R_v)$
        \STATE $\theta_{E}^{*},\theta_C^*  \stackrel{+}{\longleftarrow} - \nabla_{\theta_{E},\theta_C} (\mathrm{CE}(P, L_v))$ $\quad\triangleleft$ cross-entropy
        \ENDFOR

        \STATE \textbf{Step 2: Fine-tuning BearLLM}
        \STATE Init. $\theta_A$ by Eq. \ref{eq:get_weight}
        \FOR{$e \leftarrow 1$ to 20 epoches}
            \STATE $Y = \mathrm{BearLLM}(X_t, X_v)$
            \STATE $\theta_A^*, \theta_L^* \leftarrow \mathrm{PEFT}(Y, L_t)$
        \ENDFOR

        \RETURN $\theta_E^*, \theta_A^*, \theta_L^*$
    \end{algorithmic}

\end{algorithm}

\begin{figure}
    \centering
    \includegraphics[width=3.3in]{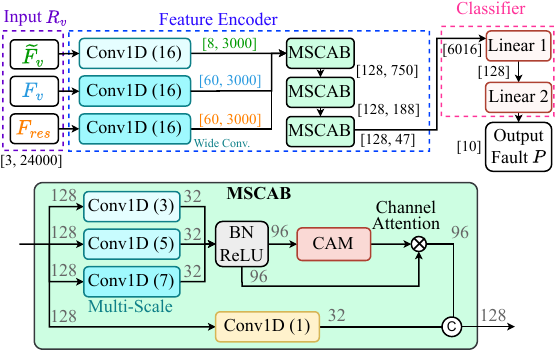}
    \caption{Structure of our proposed FCN. In the feature encoder, three wide convolutions are first used to extract main features, followed by three MSCAB blocks to transform and fuse multi-scale features for fault classification. The pre-trained FCN is used to initialize the feature extractor and alignment layer of BearLLM.}
    \label{fig:MSCAN}
        \vspace{-0.2cm}
\end{figure}

To extract the features of vibration signals, we propose a fault diagnosis network (FCN) containing a feature encoder parameterized by $\theta_E$ and a linear classification layer parameterized by $\theta_C$, as shown in Fig. \ref{fig:MSCAN}.
We extract features from the unified vibration signal representation ($R_v$) using three separate convolutional layers with large kernels \cite{WDCNN} and no weight sharing. We then transform features by three multiscale channel attention blocks (MSCAB) where the multiscale features are fused using the channel attention module (CAM) \cite{wooCBAMConvolutionalBlock2018a}. Finally, we use two linear layers for fault classification.

Our FCN takes unified representation ($R_v$) as input and outputs the fault type ($P$). 
The shape of $P$ is $[1, \gamma]$ and  $\gamma$ denotes the number of fault types.
We use cross-entropy loss for training with fault label $L_v$ as ground truth.
The training procedure is described in Algo. \ref{algo:training}. The well-trained feature encoder weights ($\theta_E^*$) of FCN are then used and frozen in BearLLM (see Fig. \ref{fig:framework}), while the classifier weights ($\theta_C^*$) of FCN are used to initialize the alignment layer.




\begin{table*}[htbp]
\small
    \centering
    \resizebox{\textwidth}{!}{
    \begin{tabular}{p{1.5cm}|l|c|c|c|c|c|c|c|c|c|c}
    \hline
        \multicolumn{2}{c|}{Method} &  CWRU &  DIRG &  HIT &  IMS &  JNU &  JUST &  MFPT &  PU &  XJTU & MBHM \\ \hline
        \multicolumn{12}{l}{Train and Test on Individual Datasets} \\ \hline
        \multirow{3}{1.5cm}{Methods for specific conditions} & WDCNN & 71.60 & 41.23 & 94.42 & 96.03 & 55.71 & 49.67 & 75.00 & 75.84 & 95.99 & - \\ 
        & TCNN & 81.56 & 39.31 & 91.03 & 91.48 & 70.37 & 48.83 & 87.50 & 72.76 & 94.36 & - \\
        & QCNN & 80.55 & 46.58 & 96.04 & 95.46 & 51.06 & 48.67 & 87.50 & 78.46 & 96.49 & - \\ \hline
        \multicolumn{12}{l}{Train on MBHM Dataset and Test on Individual Datasets} \\ \hline
        \multirow{6}{1.5cm}{Methods for specific conditions} & WDCNN & 34.22 & 25.01 & 72.56 & 93.60 & 40.20 & 41.38 & 37.50 & 56.82 & 88.98 & 65.79 \\ 
        & TCNN & 15.98 & 14.98 & 54.05 & 91.87 & 29.91 & 26.43 & 25.00 & 51.69 & 85.00 & 57.20 \\ 
        & QCNN & \colorbox{lightgray}{48.01} & \colorbox{lightgray}{27.54} & \colorbox{lightgray}{78.38} & \colorbox{lightgray}{94.23} & 43.99 & 42.56 & \colorbox{lightgray}{62.50} & \colorbox{lightgray}{63.06} & \colorbox{lightgray}{90.26} & 67.87 \\ 
        & WDCNN+DCN & 80.35 & 68.76 & 96.51 & 96.39 & 90.87 & 77.78 & 62.50 & 84.86 & 95.66 & 87.60 \\ 
        & TCNN+DCN & 48.40 & 28.19 & 68.60 & 92.41 & 46.98 & 48.19 & 12.50 & 68.49 & 90.96 & 68.98 \\ 
        & QCNN+DCN & \underline{95.55} & \underline{91.32} & \underline{99.41} & 97.36 & \underline{98.96} & \underline{85.91} & \underline{100.00} & 92.32 & \underline{98.27} & 87.60 \\ \hline
        \multirow{6}{1.5cm}{Methods for cross conditions} & MagNet & 12.07 & 15.80 & 42.50 & 91.96 & 29.09 & 27.31 & 25.00 & 30.68 & 85.23 & 55.54 \\ 
        & BearingFM & 41.18 & 14.29 & 33.33 & 92.18 & \colorbox{lightgray}{48.39} & \colorbox{lightgray}{58.78} & 33.33 & 54.22 & 88.45 & \colorbox{lightgray}{75.18} \\ 
        & MagNet+DCN & 95.31 & 79.94 & 89.78 & 88.79 & 94.95 & 65.85 & 87.50 & 74.43 & 95.85 & 81.18 \\ 
        & BearingFM+FCN & 81.16 & 20.69 & 80.00 & \underline{97.74} & 81.18 & 83.26 & 25.00 & \underline{95.31} & 95.17 & \underline{90.07} \\ 
        & \multirow{2}{1.5cm}{\textbf{Ours}} & \textbf{100.00} & \textbf{99.72} & \textbf{99.90} & \textbf{99.39} & \textbf{99.44} & \textbf{98.16} & \textbf{100.00} & \textbf{99.41} & \textbf{98.79} & \textbf{99.02} \\
        & & \textcolor{red}{(+108\%)} & \textcolor{red}{(+262\%)} & \textcolor{red}{(+28\%)} & \textcolor{red}{(+6\%)} & \textcolor{red}{(+106\%)} & \textcolor{red}{(+67\%)} & \textcolor{red}{(+60\%)} & \textcolor{red}{(+58\%)} & \textcolor{red}{(+10\%)} & \textcolor{red}{(+32\%)} \\ \hline
    \end{tabular}}
    \caption{Accuracy comparison with existing methods. ``+DCN'' denotes the addition of DCN to the original method, while ``+FCN'' indicates the replacement of the network of the original method with FCN, ``(+108\%)'' represents a relative improvement from 48.01\% to 100\%. Our approach not only surpasses the SOTA accuracy on the MBHM dataset but also achieves results superior to those obtained from models trained specifically for individual datasets. The DCN and FCN components demonstrate broad applicability across diverse scenarios.}
    \label{tab:acc_compare}
    \vspace{-0.2cm}
\end{table*}

\subsection{Feature Alignment}

We propose a feature alignment layer to embed vibration features into word embedding, which is an MLP consisting of three linear layers (i.e., $l_1, l_2, l_3$). 
The weights of alignment layer is $\theta_A=[\theta_C^{*},\theta_{l_3}]$, where $\theta_C^{*}$ is the weights of $l_1 \& l_2$ (i.e., two linear classification
layers in FCN) and $\theta_{l_3}$ is the weights of $l_3$.
We use $l_3$ transforms the output $P$ of $l_2$ into the word embedding $H_v=\mathrm{reshape}(l_3(P))$, i.e., 
\begin{equation}
    P \in \mathbb{R}^{1 \times \gamma}
    \stackrel{l_3}{\longrightarrow}
    \mathbb{R}^{1 \times \tau h}
    \stackrel{\mathrm{reshape}}{\longrightarrow}
    H_V \in \mathbb{R}^{\tau \times  h},
    \label{eq:align}
\end{equation}
where $\tau$ signifies the token length after transformed, $h$ is the hidden size of the LLM.

The weight $\theta_{l_3}$ of $l_3$ is initialized from the textual descriptions $K$ of all fault categories by
\begin{equation}
    K \in \mathbb{T}^{\gamma \times 1}
    \stackrel{\bf{T}}{\rightarrow}
    \mathbb{R}^{\gamma \times \tau}
    \stackrel{\bf{E}}{\rightarrow}
    \mathbb{R}^{\gamma \times \tau \times h}
    \stackrel{\mathrm{reshape}}{\longrightarrow}
    \theta_{l_3} \in \mathrm{R}^{\gamma \times \tau h},
    \label{eq:get_weight}
\end{equation}
where $\mathbb{T}$ stands for the text domain. $\bf{E}$ and $\bf{T}$ indicate the embedding layer and tokenizer of the pre-trained LLM, respectively. Using a tokenizer $\bf{T}$ and an embedding layer $\bf{E}$, we generate a word embedding from $K$, which is then reshaped into the weight matrix $\theta_{l_3}$. See Appendix C.3 for more details on initializing weights.

We use the pre-trained Qwen2-1.5B \cite{qwen} as our LLM parameterized by $\theta_L$, achieving basic human-computer interaction. However, its knowledge of specific domains and generation quality still requires improvement. We used the existing LoRA technique \cite{hu2021loralowrankadaptationlarge} and a general pipeline PEFT \cite{peft} for simultaneous fine-tuning of the LLM and our proposed alignment layer, which is detailed in Algo. \ref{algo:training}. 


\section{Experiments}

\subsection{Experimental Setup}



We implemented the proposed method using PyTorch \cite{PyTorch}. Both pre-training and fine-tuning are performed on a single Nvidia RTX 4090 GPU. For pre-training, comparison trials, and ablation experiments, we used AdamW \cite{AdamW} as the optimizer, and the batch size was set to 1024 for up to 50 epochs of training. Fine-tuning was performed using the existing PEFT \cite{peft} library.

To evaluate the effectiveness of our method, we provide quantitative comparison results for fault diagnosis, ablation of key components, and a user study to assess the quality of language responses. We addressed potential label leakage by dividing the 9 public datasets into a 7:2:1 ratio individualy. The training set for the MBHM dataset consists of the concatenated training sets of these individual datasets, ensuring no overlap with their corresponding test sets. Other tasks including anomaly detection, maintenance recommendations, and potential risk analysis can be found in Appendix D.

\subsection{Comparison with Fault Diagnosis Methods}

We compared BearLLM with the following fault diagnosis methods. BearFM \cite{BFM} and MagNet \cite{MagNet} are intended for diagnosing faults under cross-working conditions, while WDCNN \cite{WDCNN}, TCNN \cite{TCNN}, and QCNN \cite{QCNN} are aimed at handling specific working conditions. Detailed descriptions of these methods can be found in Appendix B. To ensure a fair comparison, we re-implement these methods and test them under the same setup in section 5.1. The results are displayed in Tab. \ref{tab:acc_compare}.

\begin{figure}[ht]
\centering
\includegraphics[width=3.3in]{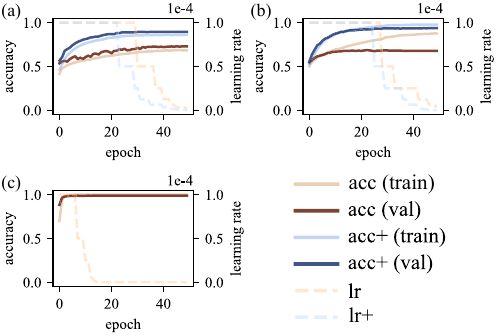}
    \caption{Accuracy and learning rate trends during training for different models. (a) Replacing the network of BearingFM with FCN resulted in increased accuracy and accelerated convergence. (b) Incorporating DCN into QCNN significantly mitigated overfit. (c) Our proposed method exhibits the fastest convergence and highest accuracy.}
    \vspace{-0.2cm}
    \label{fig:train_compare}
\end{figure}

Our DCN achieves greater accuracy compared to BearingFM \cite{BFM} when used with the same FCN (see Fig. \ref{fig:train_compare} (a)). The reason for this enhancement is likely due to BearingFM using absolute values after the FFT of the envelope spectrum. This method captures only the amplitude and ignores crucial phase information. In contrast, DCN leverages real-number computations, which help to reduce potential information loss, and operates in less than 20\% of the time required by the comparison method. Combining DCN with MagNet \cite{MagNet} and utilizing aligned data for fusion augmentation has noticeably improved performance on datasets with substantial distribution differences. 

Reflected in Tab. \ref{tab:acc_compare}, the three methods (WDCNN, TCNN, QCNN) lacking data augmentation or alignment indicate strong accuracy on some specific datasets. However, their capacity to manage massive distribution differences is restricted when trained on the MBHM dataset. Including DCN eases the marked overfitting in QCNN \cite{QCNN}, leading to a substantial improvement in validation accuracy (see Fig. \ref{fig:train_compare} (b)). Similarly, adding DCN to both WDCNN \cite{WDCNN} and TCNN \cite{TCNN} led to higher accuracy. Among all the methods tested, our proposed method achieves the highest accuracy and converges the fastest (within 20 epochs on the MBHM dataset as shown in Fig. \ref{fig:train_compare} (c)). 

\subsection{Ablation Experiments and Generalization}

\begin{table}[H]
    \centering
    \begin{tabular}{crrr}
    \hline
        $n_f$ & Param. & FLOP & Accuracy \\ \hline
        6,000 & 0.4013M & 0.0177G & 97.70\% \\ 
        12,000 & 0.5979M & 0.0353G & 98.32\% \\ 
        24,000 & 0.9747M & 0.0704G & 99.02\% \\ 
        48,000 & 1.7448M & 0.1408G & 99.20\% \\ \hline
    \end{tabular}
    \caption{Comparison of the number of parameters and FLOP of FCN under different $n_f$ settings as well as the accuracy on the MBHM dataset.}
    \label{tab:target_length}
    \vspace{-0.2cm}
\end{table}

The tests were carried out using four different $n_f$ settings in DCN (see Eq. \ref{eq:prepocess_signal}), as depicted in Tab. \ref{tab:target_length}. As vibration information primarily resides in the low-frequency range, cut is unlikely to significantly impact accuracy. By increasing the number of frequency components, the distortion due to cut can be minimized, which enhances the precision on the MBHM dataset; however, this also raises parameters and computation in FCN. To achieve a balance between accuracy and performance, we opt for 24,000 as the $n_f$.

\begin{table}[htbp]
    \centering
    \begin{tabular}{lccc}
    \hline
        \multirow{2}{1.5cm}{Method} & Accuracy & \multicolumn{2}{c}{Generalization} \\ \cline{2-4}
        ~ & MBHM & JUST & IMS \\ \hline
        Ours & 99.02\% & 90.22\% & 98.52\% \\ 
        w/o DCN & 72.15\% & 37.00\% & 91.46\% \\ 
        w/o $\tilde{F}_v, F_{res}$  & 98.35\% & 87.52\% & 97.81\% \\ 
        w/o $F_{res}$  & 98.82\% & 87.83\% & 97.96\% \\
        w/o $\tilde{F}_v$  & 98.63\% & 90.34\% & 98.29\%  \\ \hline
    \end{tabular}
        \caption{A comparison of accuracy and generalization for different ablation setups is presented.}
    \label{tab:ablation_result}
\end{table}

Ablation studies were conducted to further validate the effectiveness of each component in our proposed method. We evaluated the performance by directly using raw time-domain vibration signals (fixed-length segments) as input and removing fault-free channels and residual channels separately and together.

\begin{figure}[htbp]
\centering
\includegraphics[width=3.3in]{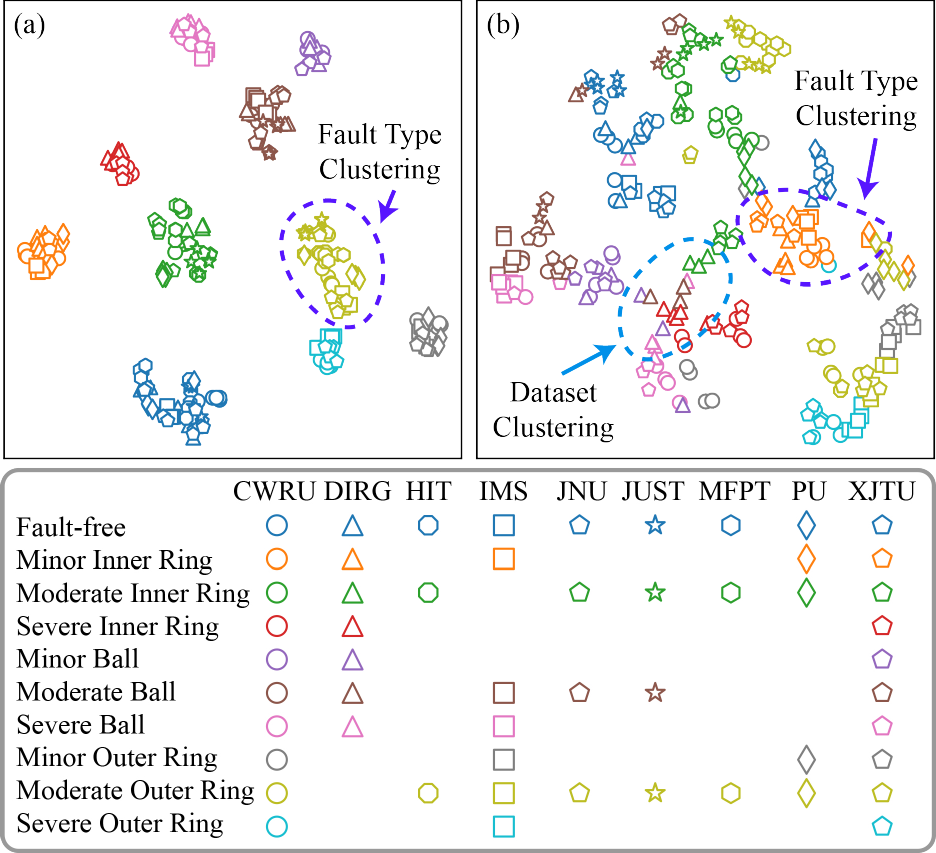}
    \caption{Visualization of output features with t-SNE. (a) Our method demonstrates clear inter-class separability. (b) Removing the fault-free and residual channels results in the signals from the same dataset exhibiting similar features.}
    \label{fig:tsne}
\end{figure}

Experimental results in Tab. \ref{tab:ablation_result} demonstrate significant accuracy and generalization drops when using time-domain signals only, further highlighting the efficacy of DCN. Applying the t-SNE, we compared the visualization of output with and without fault-free and residual channels. The blue box in Fig. \ref{fig:tsne} (b) shows how signal segments from the same dataset cluster closely in the feature space. This indicates that the model first identifies the dataset type before refining fault classification. Conversely, our proposed method, as shown in Fig. \ref{fig:tsne} (a), reduces inter-dataset differences. The model targets the residual between the query signal segments and the fault-free signal segments, creating a unified feature representation across the varying working conditions, and improving the generalization.

We evaluate the generalization ability of our proposed method using zero-shot settings. Among the publicly available datasets employed, JUST \cite{JUST} and IMS \cite{IMS} are the largest. We trained on the MBHM(w/o JUST\&IMS) dataset, comprising only 35\% of the MBHM training data, and performed zero-shot tests on the JUST and IMS datasets separately. On the JUST dataset, our method achieves an accuracy of 90.22\% without any fine-tuning. In contrast, the method without fault-free and residual channels achieves an accuracy of only 87.54\%.

\begin{figure*}[htbp]
    \centering
    \includegraphics[width=7in]{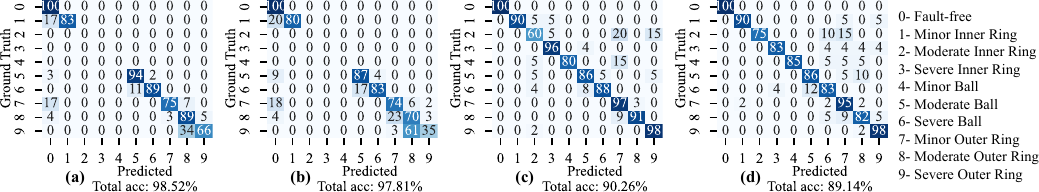}
    \caption{Confusion matrices for zero-shot performance in various scenarios. (a) Our method trained on the MBHM(w/o IMS\&JUST) and tested on the IMS shows relatively reliable accuracy. (b) The method without fault-free and residual channels trained on the MBHM(w/o IMS\&JUST) and tested on the IMS displays lower accuracy and a tendency to underestimate severity. (c) Our method trained on the MBHM(w/o CWRU) and tested on the CWRU confirms generalization. (d) Our method trained on the MBHM(w/o CWRU\&XJTU) and tested on the CWRU further verifies the generalization and efficacy of the unified representation.}
        \vspace{-0.35cm}
    \label{fig:confusion_matrix}
\end{figure*}

Fig. \ref{fig:confusion_matrix} (a,b) illustrates a comparison of confusion matrices for zero-shot testing on the IMS dataset \cite{IMS}, with and without fault-free and residual channels. Given that the IMS dataset is unbalanced (most samples are fault-free), the overall accuracy drops slightly from 98.52\% to 97.81\%. However, the method without two auxiliary channels tends to grossly underestimate the severity. For example, 61\% of severe outer ring faults are classified as moderate, and 23\% of moderate outer ring faults are identified as minor.

The CWRU \cite{CWRU1} and XJTU \cite{XJTU} datasets are the only ones that include all ten types of faults. To confirm the potential to create a unified representation, we trained our model on the MBHM(w/o CWRU) and MBHM(w/o CWRU\&XJTU) datasets, respectively. We then performed zero-shot testing on the commonly used CWRU dataset, with the results of the confusion matrices displayed in Fig. \ref{fig:confusion_matrix} (c,d). Our method achieves remarkable accuracies of 90.26\% and 89.14\% on the untrained CWRU dataset for each setting, respectively. This result is even better than some methods trained on CWRU, which shows the generalization of our unified representation method and does not depend on any specific complete dataset for training.

\subsection{User Study}

\begin{table}[htbp]
    \centering
    \begin{tabular}{lcccc}
    \hline
    Task & A & B & C & D \\ \hline
    FCN & 9\% & 10\% & - & - \\
    w/o fine-tune & 44\% & 32\% & 46\% & 45\% \\
    BearLLM & \textbf{47\%} & \textbf{58\%} & \textbf{54\%} & \textbf{55\%} \\ \hline
    \end{tabular}
    \caption{Voting results from user study.  Tasks A-D corresponds to anomaly detection, fault diagnosis, maintenance recommendations, and potential risk analysis. The fine-tuned BearLLM was the most favored across all tasks.}
    \vspace{-0.2cm}
    \label{tab:vote_result}
\end{table}

\begin{figure}[htbp]
    \centering
    \includegraphics[width=3.3in]{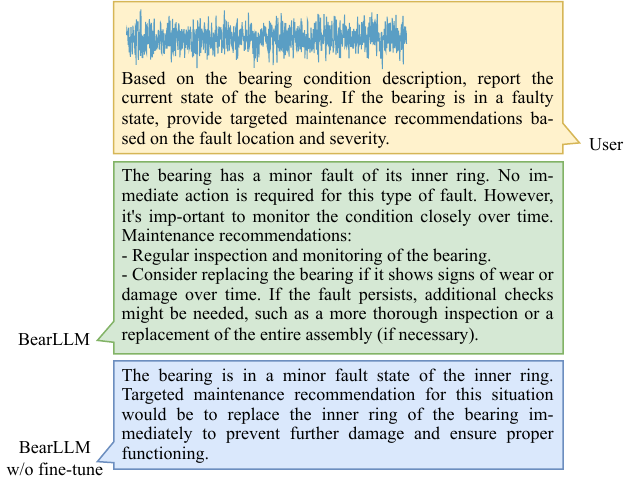}
    \caption{Examples of inputs and outputs of BearLLM. Vibration signals and task requirements are provided as user input, resulting in relevant natural language text output. The fine-tuned BearLLM exhibits improved response quality.}
    \label{fig:finetune_response}
    \vspace{-0.2cm}
\end{figure}

Tab. \ref{tab:vote_result} summarizes the outcomes of four different tasks, with users choosing the best outputs from FCN, untuned BearLLM, and fine-tuned BearLLM in blind trials. Notably, in simpler tasks, few users chose the fault code output, while most preferred the natural language output. Fig. \ref{fig:finetune_response} illustrates examples of outputs before and after fine-tuning. Appendix D provides further comparisons for various tasks. Fine-tuning did not significantly affect the output of the simple anomaly detection task. In the fault diagnosis task, the model without fine-tuning sometimes missed information on fault severity, an issue that was resolved with fine-tuning. For the two more complex tasks, the fine-tuned model produced more accurate and detailed responses. Our method addresses the challenge faced by non-experts in utilizing maintenance systems due to their complexity, reducing the required level of expertise.


\section{Conclusion}

We propose BearLLM, a novel multimodal bearing health management framework that is the first attempt to unify multiple bearing-related tasks using LLMs, including anomaly detection, fault diagnosis, maintenance recommendations, and potential risk analysis.
To build this unified framework, we introduce a prior knowledge-enhanced vibration signal representation for hundreds of different working conditions and construct the first large-scale multimodal bearing health management (MBHM) dataset.
Experimental results on nine public fault diagnosis datasets show that BearLLM outperforms state-of-the-art methods, even surpassing those specifically trained on individual datasets.
In addition, our frequency domain input alignment and feature extraction modules are plug-and-play, significantly improving the performance of other fault diagnosis models. 
We hope our work can inspire future research on building more capable industrial multimodal models.

\section{Acknowledgments}
This work was supported by National Natural Science Foundation of China under grant No. 62073312, Applied Basic Research Program of Liaoning Province (2023JH2/101300228, 2023JH2/101300143), Natural Science Foundation of Liaoning Province (2022-MS-033).

\clearpage

\bibliography{aaai25}

\clearpage
\section*{Appendix}

\subsection*{A. Construction of MBHM Dataset}

\subsubsection*{A.1. Vibration Signals}

We perform non-overlapping sampling at equal timing times on nine publicly available datasets, i.e., CWRU \cite{CWRU1}, DIRG \cite{DIRG}, HIT \cite{HIT}, IMS \cite{IMS}, JNU \cite{JNU1}, JUST \cite{JUST}, MFPT \cite{MFPT1}, PU \cite{PU}, XJTU \cite{XJTU}.

\begin{figure}[H]
    \centering
    \includegraphics[width=3.3in]{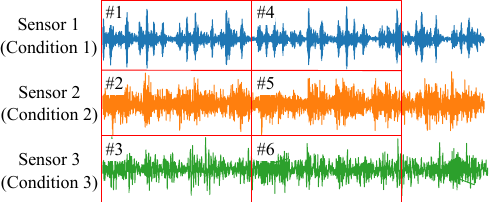}
    \caption{Sampling method for vibration signals. The vibration signals of each sensor have differences. Each signal sample has an equal duration, based on the sensor sampling rate.}
    \label{fig:sampling}
    \vspace{-0.4cm}
\end{figure}

These datasets typically involve multiple vibration sensors performing simultaneous signal acquisition. These vibration signals reflect different time-domain characteristics (see Fig. \ref{fig:sampling}) due to differences in sensor designations, mounting locations and orientations. We generalize the different sensors as part of the working conditions, i.e., the same working conditions represent the same sensors, speeds, and loads from the same dataset.

\newcommand{\cinfo}{C_{\texttt{info}}}
\newcommand{\clist}{C_{\texttt{list}}}
\newcommand{\insertarrow}{\stackrel{\textrm{insert}}{\longleftarrow}}
\newcommand{\ttrpm}{\texttt{rpm}}
\newcommand{\ttload}{\texttt{load}}
\newcommand{\ttsensor}{\texttt{sensor}}
\newcommand{\comment}{$\quad\triangleleft\quad$}

\begin{algorithm}
    \caption{Obtaining vibration signals from publicly available datasets}
    \label{algo:vibration}
    \begin{algorithmic}

    \REQUIRE $X_o, L_v, s, \ttrpm, \ttload, \ttsensor$ the raw vibration signals, fault labels, sampling rates, speeds, loads, sensor information from publicly available datasets; $D_{1-9}$ the nine publicly available datasets.
    \ENSURE MBHM(Vibration) dataset.
        \STATE $\clist=$\texttt{[ ]} \comment Initialize working conditions list
        \FOR{$D_i$ \textbf{in} $D_{1-9}$}
        \FOR{$X_o, L_v, s, \ttrpm, \ttload, \ttsensor$ \textbf{in} $D_i$}
        \STATE $\cinfo = \textbf{string}(\ttrpm, \ttload, \ttsensor, D_i)$
        \IF {$\cinfo\textbf{ not in }\clist$}
        \STATE $\clist \insertarrow \cinfo$ \comment new working condition
        \ENDIF
        \STATE $C=\textbf{find\_index}(\cinfo \textbf{ in } \clist)$
        \STATE $X_v = \textbf{sample}(X_o, s)$ \comment see Fig. \ref{fig:sampling}
        \STATE MBHM(Vibration) $\insertarrow X_v, L_v, C$
        \ENDFOR
        \ENDFOR
        \RETURN MBHM(Vibration)
    \end{algorithmic}
\end{algorithm}

We collected vibration signals $X_v$ and fault labels $L_v$ from these datasets as the vibration signal portion of the MHBM dataset (details in Algo. \ref{algo:vibration}) while abstracting the specific working condition information $\cinfo$ into condition index $C$ to facilitate quick indexing of reference vibration signals for the same working conditions.

\begin{figure}[H]
    \centering
    \includegraphics[width=3.3in]{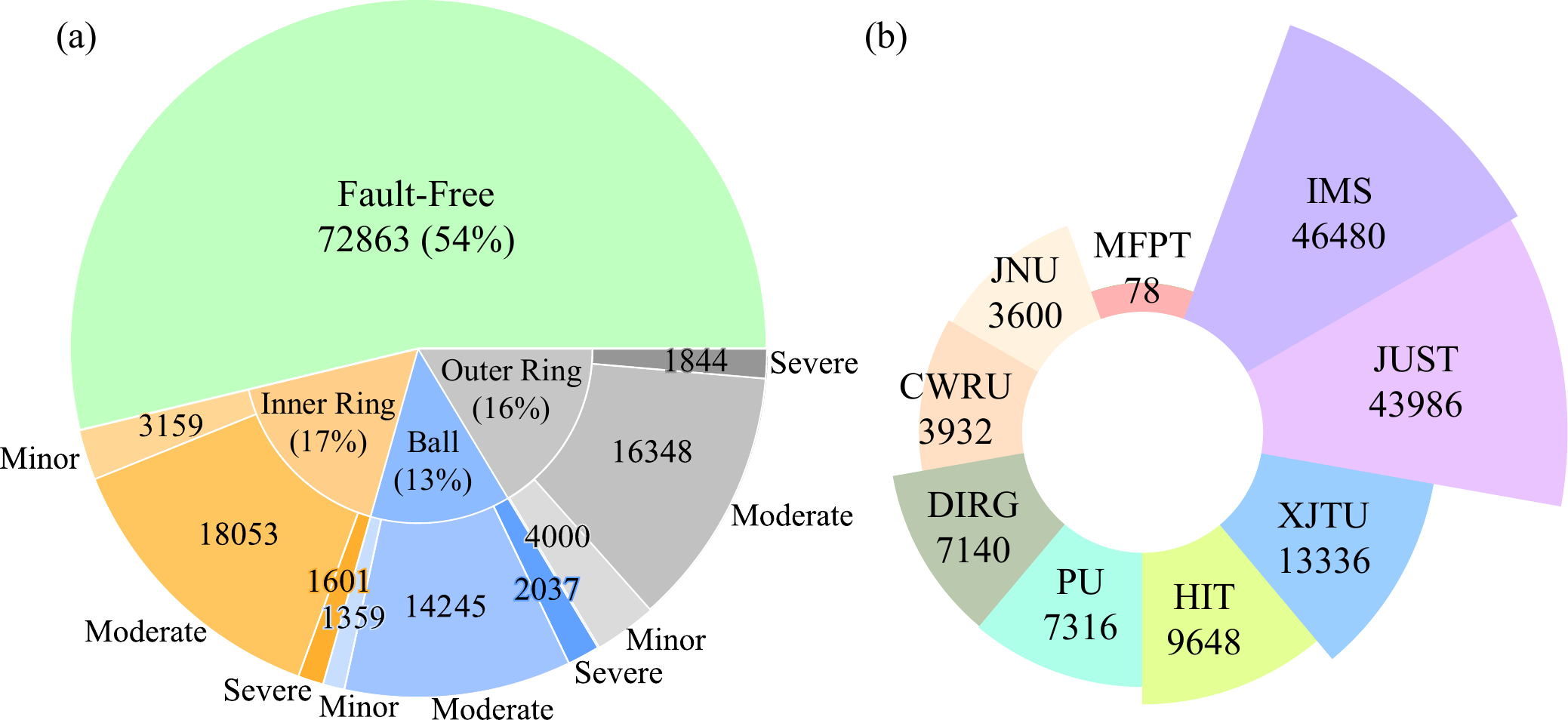}
    \caption{Fault types and data sources for the MBHM dataset. (a) MBHM contains the complete set of 10 types, (b) MBHM contains vibration signals from 9 datasets.}
    \label{fig:pie_chart}
    \vspace{-0.4cm}
\end{figure}

Figure \ref{fig:pie_chart} further illustrates the fault types and data sources for the MBHM dataset. In this case, 54\% of the samples are fault-free. The three fault locations (i.e., inner ring, ball, and outer ring) are relatively balanced. Moderate failures accounted for the majority of the three failure levels (i.e., minor, moderate, and severe). More than half of all vibration signal data came from the IMS and JUST datasets.

\subsubsection*{A.2. Generation of Text Responses}
For LLMs, the corpus consists of three parts, i.e., system prompts $X_{\texttt{sys}}$, user prompts $X_t$, and responses $L_t$. The system prompts and user prompts are taken as inputs and response text is the output of LLM. For all samples, the same system prompt text $X_{\texttt{sys}}$ is provided:

\texttt{\small As an expert in bearing fault diagnosis with extensive knowledge in mechanical engineering and failure analysis, you can assess the condition of bearings. Typically, bearing states are categorized as normal, outer ring fault, inner ring fault, and ball fault. These defects are further classified into three levels: minor, moderate, and severe. Based on your description of the bearing state, you will answer my questions concisely and directly, providing only the answer without reiterating the user's prompt or bearing status description.}

We provide templates $\tilde{X}_t$ for user prompts $X_t$ for each of the four different types of tasks:

\begin{itemize}
    \item \textbf{Anomaly Detection:} \texttt{\small Bearing status descript-
    ion: \#placeholder\#. Based on the bearing condition description, determine whether the bearing is in a faulty state. Answer yes or no.}
    \item  \textbf{Fault Diagnosis: }\texttt{\small Bearing status description: \#placeholder\#. Based on the bearing condition description, identify the type of bearing fault. Bearing conditions are classified as normal, outer ring fault, inner ring fault, and ball fault. All defects are categorized into three levels: minor, moderate, and severe.}
    \item \textbf{Maintenance Recommendations: }\texttt{\small Bearing status description: \#placeholder\#. Based on the bearing condition description, report the current state of the bearing. If the bearing is in a faulty state, provide targeted maintenance recommendations based on the fault location and severity.}
    \item \textbf{Potential Risk Analysis }\texttt{\small Bearing status descr-
    iption: \#placeholder\#. Based on the bearing condition description, assess the potential risks associated with the bearing condition. Identify the potential consequences of the bearing fault and recommend appropriate actions to prevent catastrophic failures.}
\end{itemize}

The template $\tilde{X}_t$ is embellished and modified using LLMs to simulate different user inputs $X_t$ while maintaining the meaning and not modifying the placeholders. We use the leading ChatGPT \cite{openaiGPT4TechnicalReport2024} for response corpus generation. In the generation, \texttt{\small\#placeholder\#} is replaced with fault descriptions, according to fault labels $L_v$, e.g., \texttt{\small moderate fault of bearing outer ring}. 
We simulated four separate tasks for each sample in the MBHM(Vibration) dataset to generate the MBHM dataset, details of which are given in Algo. \ref{algo:mbhm}.

\begin{algorithm}
    \caption{Algorithm for building the MBHM dataset}
    \label{algo:mbhm}
    \begin{algorithmic}
    \REQUIRE $X_v, L_v, C$ vibration signal, fault label, working condition from MBHM(Vibration) dataset, $T_{A-D}$ task types, $X_{\texttt{sys}}$ system prompt, $\tilde{X}_t$ user prompt templates.
    \ENSURE MBHM dataset.
        \FOR{$T \textbf{ in } T_{A-D}$}
        \STATE $X_t = \textbf{mod}(\tilde{X}_t | _T$) \comment simulate user inputs
        \STATE $L_t = \textbf{ChatGPT}(X_{\texttt{sys}},X_t,L_v)$
        \STATE MBHM $\insertarrow X_v, L_v, C, X_t, L_t$        
        \ENDFOR
        \RETURN MBHM
    \end{algorithmic}
\end{algorithm}

\subsection*{B. Differences between Ours and Comparative Methods}

\subsubsection*{B.1. Methods to Cope with Single Working Condition}
The main existing fault diagnosis methods are designed for a single working condition only, and we select several representative methods for comparison.
The \textbf{WDCNN} \cite{WDCNN} is arguably the most popular diagnostic network, incorporating BatchNorm for fault diagnosis and demonstrating the effectiveness of using larger kernels in the first convolutional layer for improved accuracy. Due to its straightforward architecture, it enjoys widespread application in both practical scenarios and methodological comparisons.
The \textbf{TCNN} \cite{TCNN} presents a potential enhancement by adding Dropout techniques and increasing the depth of the network to augment feature learning capabilities from raw data. The \textbf{QCNN} \cite{QCNN} introduced quadratic convolution to the fault diagnosis domain, improving diagnostic accuracy through enhanced non-linear representational ability within convolutional layers. 
In contrast to our methods, all of these methods utilize raw vibration signals as input.

\subsubsection*{B.2. MagNet with Data Augmentation}

The \textbf{MagNet} \cite{MagNet} enhanced the mixup data augmentation method, transitioning from a Beta distribution (mixing two distributions) to a Dirichlet distribution (mixing multiple distributions).
During training, in addition to a classification head, a discriminator was designed via adversarial training to render the obtained features difficult for correct source domain identification. This process compelled the feature extractor to learn common features across domains. The authors also introduced a self-adaptive screening weight strategy to mitigate the use of feature-deficient samples in the augmentation sample synthesis. 

Similar to our method, this approach attempts to transform the vibration signal from multiple independent distributions into a smooth single distribution.
However, our approach achieves alignment through simple spectral changes, whereas MagNet performs signal mixing in the time domain, which made it difficult to perform effective sample mixing in cases with large distributional differences such as our MBHM dataset.

\subsubsection*{B.3. BearingFM with Data Preprocessing}

The \textbf{BearingFM} \cite{BFM} employs a resampling strategy to align input signals to the angular domain. This method assumes that the bearing's rotational speed and sampling frequency are known, enabling resampling of the raw signal to a uniform target speed and sampling rate. Subsequently, it utilizes the Hilbert transform and FFT to extract the envelope spectrum of the signal. Finally, further data augmentation is performed through translation and scaling operations on the signal in both the frequency and amplitude axis to model input.

The similarity to our method lies in the use of preprocessing techniques to uniformly represent signals with different sampling rates. However, BearingFM requires more a priori knowledge (the RPM value of the test rig is needed) and performs more complex calculations. What's more, the authors take absolute values after the FFT, resulting in a loss of phase information of the vibration signal.

\subsection*{C. Details of Experience}

\subsubsection*{C.1. Details of Experimental Setup}

All training and testing were conducted on a Windows 11 system equipped with a Core i7-13700F CPU and a single RTX 4090 GPU. Python and PyTorch \cite{PyTorch} versions utilized were 3.11 and 2.3.1, respectively.

A batch size of 1024 was employed for pre-training, comparison trials, and ablation experiments, with an initial learning rate of $10^{-4}$. AdamW \cite{AdamW} served as the optimizer, and the learning rate scheduler was set to ReduceLROnPlateau with parameters patience=150 and factor=0.5. This implies that if the loss did not decrease for consecutive 150 batches, the learning rate would be halved. A maximum of 50 epochs was allowed, and training was considered converged and terminated prematurely if the learning rate fell below $10^{-7}$. Fine-tuning was performed using the existing PEFT \cite{peft} library.

\subsubsection*{C.2. Pre-training}

We use the MBHM dataset to pre-train the Fault Classification Network (FCN). To prevent the problem of data leakage, we first randomly divide the data into training, validation, and testing sets in the ratio of 7:2:1, and subsequent query reference signal operations are performed only within the training set. The training set is used to optimize the FCN weights, the validation set is used to evaluate the degree of overfitting of the training accuracy, and the test set is loaded with the validation set weights with the highest accuracy to evaluate the overall accuracy of the model.

\subsubsection*{C.3. Initialize weights}

\begin{figure}[htbp]
    \centering
    \includegraphics[width=3.3in]{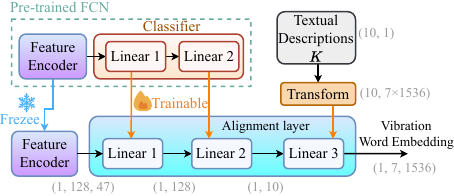}
    \caption{Initialization methods for feature encoder and alignment layer in BearLLM. The pre-trained FCN provides the feature encoder and the weights of $L1$ and $L2$. The weights of $L3$ are obtained from the transformation of the fault text description.}
    \label{fig:init_weight}
    \vspace{-0.4cm}
\end{figure}

\begin{figure}[htbp]
    \centering
    \includegraphics[width=3.3in]{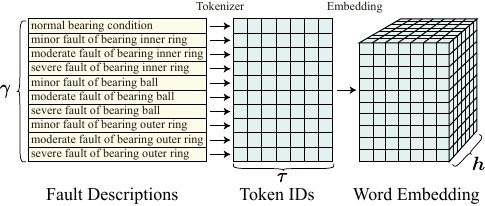}
    \caption{Initialization method for linear layer 3 weights. Firstly, each fault category is described in the text, and then word embedding is obtained as weights by a pre-trained tokenizer and embedding.}
    \label{fig:l3_weight}
    \vspace{-0.2cm}
\end{figure}

Using the pre-trained FCN and text descriptions for each fault type, the feature encoder and alignment layer of BearLLM are initialized, as shown in Fig. \ref{fig:init_weight}. The feature encoder weights are frozen and not involved in fine-tuning, and the alignment layer parameters are trainable. The specific implementation of converting fault text descriptions into weights of $L3$ in the alignment layer is shown in Fig. \ref{fig:l3_weight}.

\subsubsection*{C.4. Fine-tuning}

We utilize the LoRA technique \cite{hu2021loralowrankadaptationlarge} to build LoRA adapters for all linear layers in BearLLM, i.e., the alignment layer and the LLM.
We modify the embedding of Qwen2 \cite{qwen} to automatically replace the text embedding of the \texttt{\small\#placeholder\#} in $H_t$ with a vibration word embedding $H_v$ that is encoded and aligned by $X_v$. The replaced word embedding is fed into Qwen2 to get the output.
We use a generic PEFT pipeline \cite{peft} to calculate the difference between the output $Y$ of BearLLM and the provided $L_t$, and update the parameters of the LoRA adapters.

\subsection*{D. More Experimental Results}

\subsubsection*{D.1. Selection of $n_f$}

\begin{figure}[htbp]
    \centering
    \includegraphics[width=3in]{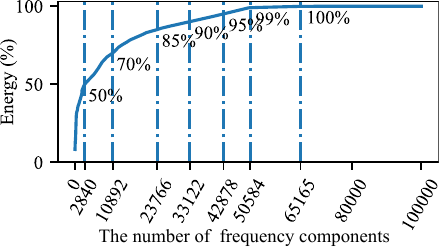}
    \caption{The DCN results contain the percentage change in the energy of the original signal at different $n_f$, by analyzing our MBHM dataset.}
    \label{fig:target_count}
    \vspace{-0.2cm}
\end{figure}

Signals with different sampling rates are converted to aligned representations by the DCN, but the number $n_f$ of frequency components in DCN needs to be specified manually, to normalize the signals to the same input length by pad or cut. We statistically analyze our MBHM dataset by evaluating the proportion of energy of the aligned signal containing the original signal for different $n_f$. The results are shown in Fig. \ref{fig:target_count}. The vibration information is mainly concentrated in the low-frequency region, and when $n_f$ reaches a certain value, continuing to increase does not lead to significant distortion improvement.

\subsubsection*{D.2. False Alarm and Missed Alarm Rates}

\begin{table}
    \centering
    \begin{tabular}{lp{2cm}p{2cm}}
    \hline
        ~ & False Alarm Rate (\%) & Missed Alarm Rate (\%) \\ \hline
        WDCNN & 24.8 & 9.4 \\ 
        TCNN & 20.3 & 22.5 \\ 
        QCNN & 23.9 & 8.3 \\
        WDCNN+DCN & 8.9 & 3.5 \\
        TCNN+DCN & 22.9 & 8.1 \\
        QCNN+DCN & 8.4 & 4 \\ 
        MagNet & 13.6 & 30.9 \\
        BearingFM & 16.9 & 7.9 \\
        MagNet+DCN & 15.7 & 3.2 \\
        BearingFM+FCN & 7.4 & 2.5 \\
        Ours (DCN+FCN) & 0.7 & 0.3 \\ \hline
    \end{tabular}
    \caption{The false alarm rate and the missed alarm rate of different methods on the MBHM dataset.}
    \label{tab:false_missed_rate}
\end{table}

 Excepte the accuracy, we have also evaluated both the false alarm rate and the missed alarm rate of all the methods on the MBHM dataset, as they are crucial metrics for fault diagnosis systems. As shown in Tab. \ref{tab:false_missed_rate}, our method shows superior performance on both metrics as well.

\subsubsection*{D.3. Effectiveness of DCN for Alignment}

\begin{figure}[htbp]
    \centering
    \includegraphics[width=3.3in]{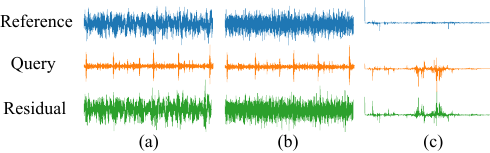}
    \caption{The residuals of the reference and query signals are calculated using different ways. (a) direct subtraction in the time domain, (b) subtraction after downsampling to the same sampling rate, (c) subtraction after alignment via DCN.}
    \label{fig:compare_residual}
    \vspace{-0.4cm}
\end{figure}

We verified the validity of the DCN alignment by calculating the residuals, as shown in Fig. \ref{fig:compare_residual}. We select a pair of reference and query (moderate inner ring fault) signals under the same working condition. The reference signal is sampled at 48 kHz and the query signal is sampled at 12 kHz. Fig. \ref{fig:compare_residual} (a) illustrates the residuals obtained by direct subtraction. Fig. \ref{fig:compare_residual} (b) illustrates the residuals obtained by phase subtraction after downsampling the reference signal to 12kHz. Both of them are difficult to reflect the difference between the query signal and the reference signal. Fig. \ref{fig:compare_residual} (c) demonstrates the residuals obtained by phase subtraction after DCN, reflecting the changes of different frequency components after a fault occurs, independent of the sampling rate difference.

\begin{figure*}[htbp]
    \centering
    \includegraphics[width=6.5in]{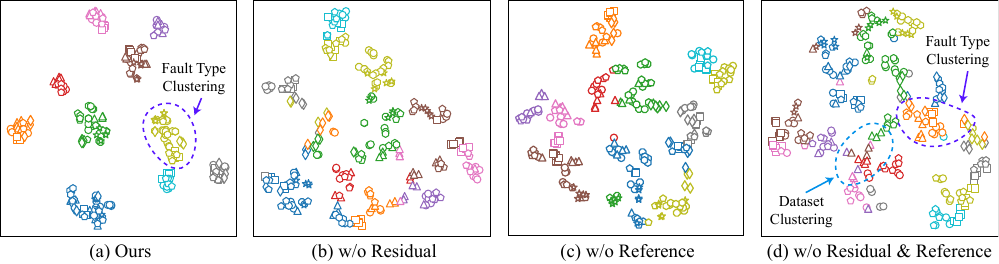}
    \caption{Visualization of t-SNE tested after training on the MBHM dataset under different ablation settings.}
    \label{fig:t_sne_more}
    \vspace{-0.3cm}
\end{figure*}

\begin{figure*}[htbp]
    \centering
    \includegraphics[width=7in]{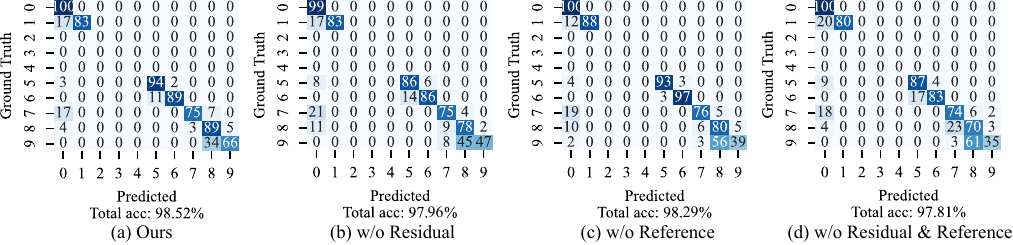}
    \caption{Confusion matrices for test accuracy on the IMS dataset trained on the MBHM (w/o IMS\&JUST) dataset at different ablation settings.}
    \label{fig:confusion_mat_more}
    \vspace{-0.3cm}
\end{figure*}

\begin{figure}[H]
    \centering
    \includegraphics[width=3.3in]{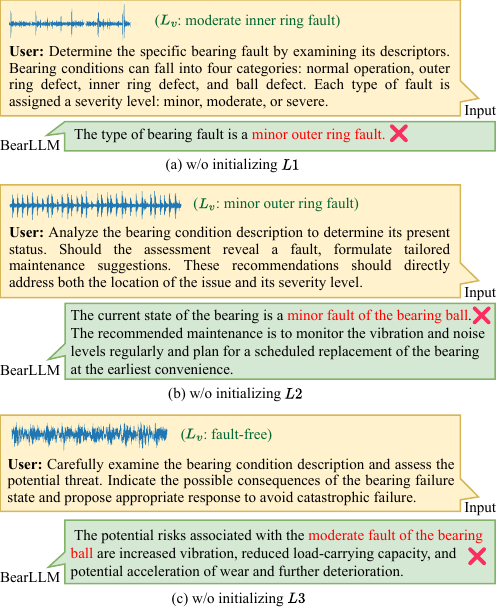}
    \caption{Missing initialization of the three linear layers of the alignment layer can lead to ineffective identification of the bearing vibration signal state, even after fine-tuning.}
    \label{fig:wo_init}
    \vspace{-0.3cm}
\end{figure}

\subsubsection*{D.4. Impact of Reference and Residual Channels}

We complement the experimental results without the residual channel $F_{res}$ and reference channel $\tilde{F}_v$. As shown in Fig. \ref{fig:t_sne_more}, without these auxiliary channels, the fault classification accuracy decreases, coming from the fact that the signals from different sources are not converted into a uniform representation, retaining some of the features of the dataset source and potentially reducing the generalizability. This is further demonstrated by 0-shot experiments on the IMS dataset \cite{IMS} in  Fig. \ref{fig:confusion_mat_more}. The confusion matrix suggests that the absence of these auxiliary channels makes the model underestimate the severity of faults, in untrained conditions. Among them, the absence of a residual channel $F_{res}$ has a greater impact on the generalizability, illustrating the validity of our efforts to build a unified representation of the vibration signal through residuals.

\subsubsection*{D.5. Impact of Initializing the Alignment Layer}

We experimentally verified the necessity of initializing the alignment layer using FCN and fault descriptions by removing the initialization steps for $L1$, $L2$, and $L3$, respectively, and implementing fine-tuning with the same settings. The results of the experiments are shown in Fig. \ref{fig:wo_init}, where the BearLLM lacking initialization can only learn the reply mode of the health management tasks, but cannot provide reliable bearing faults based on vibration signals.

\subsubsection*{D.6. More Examples of Comparing Responses Before and After Fine-tuning}

\begin{figure*}[htbp]
    \centering
    \includegraphics[width=6.5in]{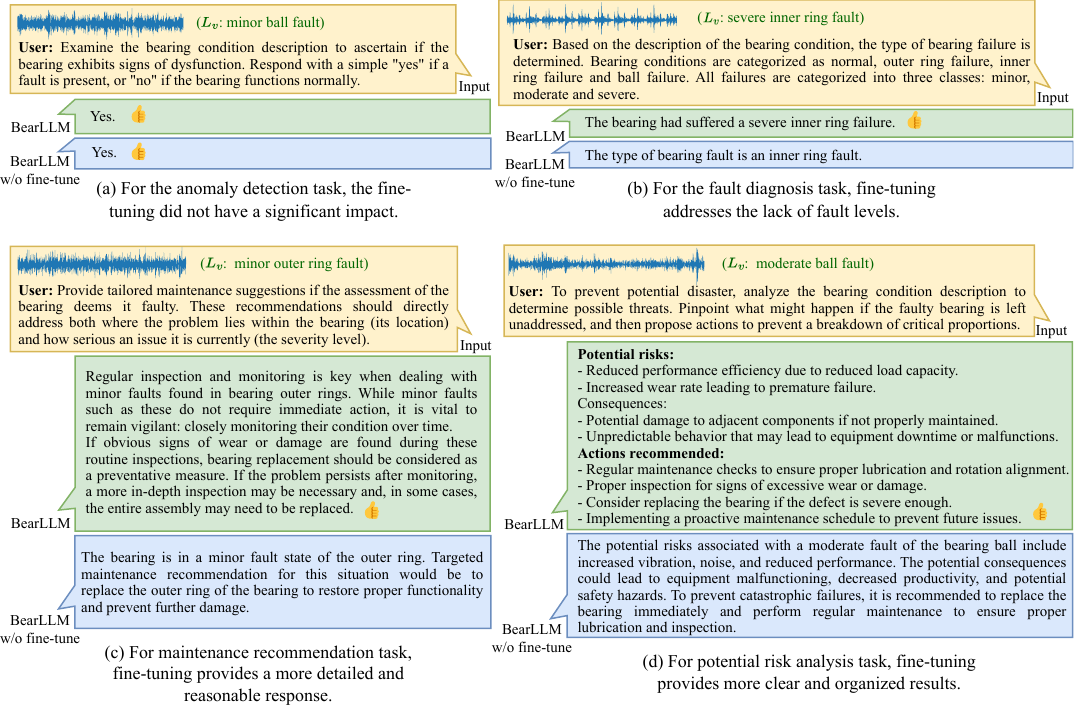}
    \caption{More response comparisons on four health management tasks, before and after fine-tuning.}
    \label{fig:more_response}
    \vspace{-0.4cm}
\end{figure*}

As shown in Fig. \ref{fig:more_response}, we demonstrate a comparison of more responses before and after fine-tuning in the four bearing health management tasks, further illustrating that fine-tuning effectively improves the ability and quality of the responses generated based on user prompt and vibration signals. High-quality responses on specific tasks can also be achieved by using smaller models through fine-tuning, which reduces the computational burden and improves the generation speed.

\subsection*{E. Limitation and Future Works}

Our approach builds a unified representation of bearing vibration signals based on a priori knowledge enhancement. However, it relies on comparisons via fault-free signals under the same working conditions, which suggests that it cannot handle bearing health management tasks without prior fault-free knowledge. For example, during the equipment acceptance process, no fault-free history exists, only when the equipment acceptance is complete, our method can be used. At the same time, signal comparison requires the same working conditions, so it cannot be applied to devices such as industrial robotic arms, whose working conditions (e.g., speed, load) vary over time because it is difficult to query the same working conditions.

In future work, more bearing health management tasks, such as remaining useful life prediction, can be carried out using the unified representation of vibration signals we have established. This frequency domain transformation can also be extended to more rotating mechanical components such as gears. Our current work utilized ChatGPT to generate text of four simulated health management tasks, and other relevant information provided in the dataset, such as test rig descriptions and bearing designations, could be incorporated into the dataset in the future to generate more targeted text responses. Our current dataset only covers vibration signals and text, and future inputs such as infrared images, currents, torques, and other modalities will further expand the scenarios for use.

Although our method has strong generalization ability and demonstrates high 0-shot accuracy after training on the MBHM dataset, future research related to continuous learning can be conducted to improve the accuracy on unseen datasets. For systems with time-varying working conditions, a series of similar working condition signals, rather than a single sample under the same working condition, can be used as reference inputs in the future.

\clearpage

\end{document}